\documentclass{article}

\usepackage{iclr2027_conference,times}
\usepackage{graphicx}
\usepackage{multirow}
\usepackage{amsmath,amssymb,amsfonts}
\usepackage{amsthm}
\usepackage{mathrsfs}
\usepackage{xcolor}
\usepackage{textcomp}
\usepackage{booktabs}
\usepackage{algorithm}
\usepackage{algorithmicx}
\usepackage{algpseudocode}
\usepackage{listings}
\usepackage{hyperref}
\usepackage{url}
\usepackage{comment}
\usepackage{wrapfig}

\theoremstyle{definition}

\theoremstyle{remark}

\makeatletter
\def\@oddhead{%
  \vbox{%
    \hbox to \textwidth{\small \hfill \hfill}%
    \vskip 4pt
    \hrule height 0.5pt width \textwidth
  }%
}
\def\@evenhead{\@oddhead}
\makeatother

\title{Double-Take: Improving Capacity, Binding, and Selectivity in Long-Context within Bidirectional Memory}

\author{
\textbf{Vakul Raghavan}\textsuperscript{1,*},
\textbf{Utkarsh Grover}\textsuperscript{2,*},
\textbf{Mingyang Mao}\textsuperscript{2},
\textbf{Chengwen Wang}\textsuperscript{1},
\textbf{Xiaomin Lin}\textsuperscript{2}
\\[0.6em]
\small
\textsuperscript{1}Department of Mathematics and Computer Science, Rutgers University, Newark, NJ, USA
\\
\textsuperscript{2}Embodied Robotics and Automation Lab, University of South Florida, Tampa, FL, USA
\\[0.4em]
 \textsuperscript{*}Equal contribution.
 \\
\footnotesize
\texttt{vnr18@scarletmail.rutgers.edu, xlin2@usf.edu}
}

\iclrfinalcopy

\begin{document}

\maketitle

\begin{abstract}
Processing a long sequence does not guarantee that distant information remains usable. As context grows, a fixed-size memory must retain many associations, connect information across positions, and preserve relevant structure without
being overwhelmed by everything it observes. These are distinct requirements: a memory may store more information without representing the relationships needed for retrieval, and it may transport those relationships without preserving the ones that matter. Thus, the challenge is not simply extending context length, but constructing a memory that remains usable as context grows. We study this problem in large-chunk test-time training (TTT) and identify two basic failures: single-step writes accumulate interference, limiting \emph{capacity}, while self-associative writes cannot form cross-position \emph{bindings}. We introduce \textbf{Double-Take (DT)}, which repairs these failures with multi-step unpinned updates and cross-position mixing, and adds bidirectional change-sensitive writes for \emph{selective retention}. Capacity alone leaves end-to-end recall at the $0.10$ chance floor; adding binding raises it to $0.92$. Trained only on $32\,\mathrm{kb}$ genomic contexts, DT reaches $0.927$ AUROC at $131\,\mathrm{kb}$ and $0.856$ near $1\,\mathrm{Mb}$, while transferring zero-shot from human to mouse. Together, these results identify \textbf{capacity, binding, and selectivity} as three requirements for usable long-range memory.
\end{abstract}

\vspace{-0.3cm}
\section{Introduction}
\label{sec:intro}
\vspace{-0.2cm}
Long-context modeling is increasingly measured by how many tokens a model can process, but a large context window does not guarantee that the information inside it remains usable. Long-context language models can lose access to relevant information depending on where it appears in the sequence \citep{liu2024lost}, and controlled evaluations show that performance often degrades sharply as context length and retrieval complexity increase \citep{hsieh2024ruler}. Similar limitations appear in efficient sequence models, where associative recall accounts for a substantial part of the gap with attention-based models \citep{arora2024zoology}. Together, these results point to a distinction that context length alone does not capture: \emph{processing distant information and remembering it are different
problems}.

This distinction becomes more important as models scale to sequences that cannot remain explicitly addressable. Full attention preserves direct token-to-token interaction but incurs quadratic cost \citep{vaswani}; efficient architectures instead summarize or compress history through structured sequence operators or recurrent states
\citep{mamba,hyenadna,caduceus}. Once information is compressed, three
questions become unavoidable. How many distinct associations can survive as new information arrives? Can the memory preserve a relationship between events that occurred at different positions? And, when the context becomes very large, which of those relationships should remain prominent? Genomics makes all three questions concrete: individual nucleotides can matter, their regulatory partners may lie tens of kilobases to megabases away, and only a small fraction of the intervening sequence may be relevant to that interaction
\citep{rao2014,dixon2012,deepsea}. The difficulty is therefore not merely reaching farther, but preserving useful structure while doing so.

Test-time training (TTT) \citep{ttt} offers a natural way to revisit this problem because it replaces a fixed recurrent state with an adaptive fast-weight memory that is updated as the sequence is processed. Large-chunk TTT (LaCT) extends this idea efficiently to long sequences \citep{lact}. Yet examining what happens when a large chunk is written into that memory exposes the same three questions in a particularly simple form. A single update superposes many associations, so new writes interfere with those already stored. Even if that interference is reduced, same-position key--value writes contain no explicit relationship between different positions. And even if both problems are resolved, transporting every association equally gives the memory no mechanism for distinguishing structure that matters from content that merely occurred. These are three different limitations: \emph{capacity}, \emph{binding}, and
\emph{selectivity}.

This observation leads directly to \textbf{Double-Take (DT)}. Multi-step unpinned updates allow competing associations to be resolved rather than collapsed into a single write, increasing \emph{capacity}. Local cross-position mixing exposes relationships before they enter memory, enabling \emph{binding}. Finally, bidirectional change-sensitive writes expose differences between opposing contexts, providing a mechanism for \emph{selective retention}. DT therefore follows from a simple view of long-range memory: additional context becomes useful only when a model can \emph{retain enough information, relate information across positions, and preserve the relationships that matter}. Building on this mechanism we present the following contributions:

\begin{itemize}
    \item 
    We identify interference-limited capacity and the inability of self-associative writes to form cross-position bindings as two limitations of large-chunk test-time memory.

    \item 
    We repair capacity with multi-step unpinned updates, enable cross-position binding through local mixing, and introduce bidirectional memory for selective retention.

    \item 
    We show that DT remains informative more than \(30\times\) beyond its
    training context, directly recruits information outside the local receptive field through its test-time memory, and transfers zero-shot from human to mouse.
\end{itemize}
    \vspace{-0.3cm}
\section{Background}
\label{sec:background}
\vspace{-0.2cm}
\subsection{Fast-Weight Memory and Test-Time Training}
\vspace{-0.1cm}
Efficient long-context models cannot keep every past token explicitly
available, so they compress the growing sequence into a smaller state that is carried forward. This makes long sequences practical, but recall does not automatically grow with context. Associative recall remains a major source of error in efficient sequence models~\citep{zoology}, state size trades directly against recall ability~\citep{arora2024}, and fixed-size latent states face fundamental limitations on copying and retrieval ~\citep{jelassi2024repeat}. Related work has therefore explored compressed, external, and persistent memories as ways to preserve information beyond a standard recurrent state~\citep{rae2019compressive,wu2022memorizing}.

Fast-weight memory takes a different approach: rather than storing the past only in activations, it writes temporary information into parameters that change as the sequence is processed~\citep{ba2016fastweights}. Linear attention can itself be interpreted as such a fast-weight memory, where key-value
associations are written into a finite state~\citep{schlag2021linear}. Later delta-rule methods showed that changing the write rule can improve how these associations are corrected and retrieved
~\citep{schlag2021linear,yang2024parallelizing,yang2025gated}. Test-time training (TTT) extends this idea by making the hidden state a small learnable model whose parameters are updated directly from the incoming sequence~\citep{ttt}. Given keys \(K\) and values \(V\), a memory write takes the form
\begin{equation}
    W \leftarrow W - \eta \nabla_W
    \mathcal{L}\!\left(f_W(K),V\right),
\end{equation}
and a later query \(Q\) reads from the updated memory through \(f_W(Q)\).

Large-Chunk Test-Time Training (LaCT) makes this process practical at long sequence lengths by updating the fast-weight memory over large blocks of tokens ~\citep{lact}. This greatly improves efficiency, but it also makes each update responsible for many associations at once. That immediately raises three questions. Can many associations be written without interfering with one another? Can information from different positions be connected inside the memory? And, as more context arrives, which of those relationships should remain important? We refer to these as \emph{capacity}, \emph{binding}, and \emph{selectivity}. Prior fast-weight and gated-memory work already shows that both the form of the write and the control of memory updates matter for long-range retrieval~\citep{schlag2021linear,yang2025gated,behrouz2026titans};
our focus is on understanding how these three requirements arise specifically when large chunks are written into test-time memory.

\vspace{-0.1cm}
\subsection{Long-Context Genomic Modeling}
\vspace{-0.1cm}
Genomics makes these requirements especially concrete. Regulatory information can depend on individual nucleotides while interactions span tens to hundreds of kilobases through chromatin contacts and domain organization ~\citep{dixon2012,rao2014}. Akita showed that DNA sequence alone can predict three-dimensional genome folding~\citep{akita}, while Enformer demonstrated that integrating long-range sequence improves regulatory prediction ~\citep{enformer}. More recent models extend this scale further: Borzoi models regulatory activity over large genomic windows~\citep{linder2025predicting}, while HyenaDNA and Caduceus bring long-context modeling to nucleotide resolution ~\citep{hyenadna,caduceus}. These models make distant sequence increasingly accessible; the remaining question is whether a memory can preserve enough of that information, connect the relevant positions, and retain the structure that matters.
This is the question we take up next. We examine the large-chunk fast-weight write itself and ask what prevents \emph{capacity}, \emph{binding}, and \emph{selectivity} from being achieved.

\vspace{-0.3cm}
\section{Double-Take}
\label{sec:method}
\vspace{-0.2cm}
Double-Take (DT) is a fast-weight memory that processes the sequence in
chunks while carrying a persistent memory state across them. DT departs from LaCT in three ways: multiple partial inner steps with free retention instead of per-chunk weight renormalization, causal cross-position mixing on the Q/K/V streams, and a bidirectional change-point write target. For each chunk \(m\), let \(h_i\) denote the hidden representation at position \(i\), for \(i=1,\ldots,N\). From each \(h_i\), the model forms three representations: a query used to read from memory, a key used to address a memory write, and a value containing the information to be written. Specifically,

\[
q_i=P_Q h_i,\qquad
k_i=P_K h_i,\qquad
v_i=P_V h_i,
\]

where \(P_Q\), \(P_K\), and \(P_V\) are learned projection matrices.

The memory itself is a small neural network with fast-changing parameters \(W=\{W_0,W_1,W_2\}\):

\[
f_W(x)
=
W_1\!\left(
\mathrm{SiLU}(W_0x)\odot W_2x
\right).
\]

The projection matrices are learned during pretraining and remain fixed at test time, while the fast-weight state \(W\) is updated as each chunk is processed. The memory is multi-headed, with an independent fast-weight state for each head and \(d_k=d_v=d_h=32\).

\begin{figure}[ht]
        \centering
    
        \includegraphics[width=1.0\textwidth]{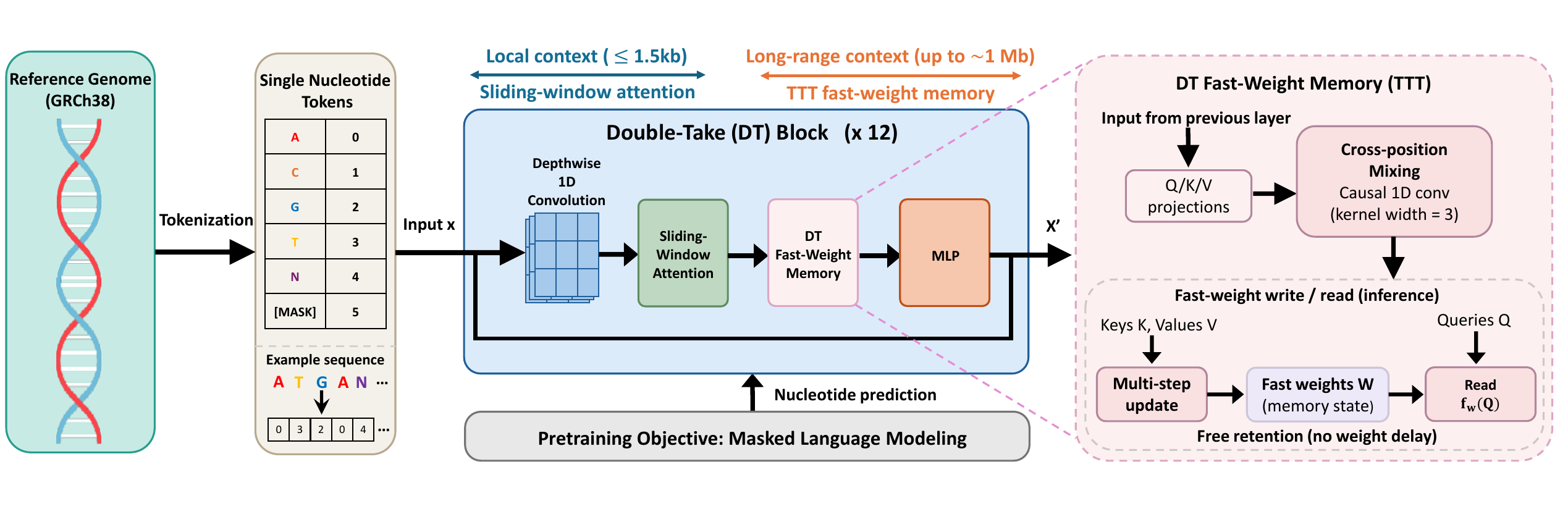}
        \caption{\textbf{Double-Take architecture.}
        Each block combines depthwise local convolution, sliding-window attention, and a persistent fast-weight memory. Local attention handles nearby interactions, while DT carries information across chunks through a fixed-size memory updated during the forward pass.}
        \label{fig:overview}
        \vspace{-0.4cm}
    \end{figure}

DT changes how information is written into this memory in three steps.
Multiple partial updates allow more associations to be stored without
overwriting the state already carried forward. Local mixing exposes
relationships between nearby positions before they are written. Finally, bidirectional views expose changes between opposing contexts so that such structure can be retained preferentially. These three mechanisms address \emph{capacity}, \emph{binding}, and \emph{selectivity}, respectively. Figure~\ref{fig:overview} summarizes the architecture.

\vspace{-0.1cm}
\subsection{Partial Updates Increase Capacity}
\label{subsec:DT-capacity}
\vspace{-0.1cm}
A large chunk writes many key-value pairs into the same fast-weight state. With a single update, these associations are written simultaneously. In the linear case \(f_M(x)=Mx\), \(\Delta M \propto \sum_i v_i k_i^\top\), so querying \(k_j\) retrieves its desired value together with contributions from other keys that overlap with
\(k_j\). As the number of pairs grows, this cross-talk increases and the one-step write becomes capacity-limited.
DT therefore does not stop after the first write. Let \(M_{m-1}\) denote the memory entering chunk \(m\). Starting from \(M_m^{(0)}=M_{m-1}\), DT performs a small number of updates,
\begin{equation}
M_m^{(s+1)}
=
M_m^{(s)}
-
\eta_s\,
\operatorname{NS}\!\left(
\nabla_M \mathcal{L}_{\mathrm{mem}}
(M_m^{(s)};K,V)
\right),
\qquad s=0,\ldots,S-1,
\label{eq:DT-update}
\end{equation}
with
\(\mathcal{L}_{\mathrm{mem}}
=-\sum_i\langle f_M(k_i),v_i\rangle\).
Here, \(\eta_s\) controls the step size and \(\operatorname{NS}(\cdot)\) denotes Muon's Newton-Schulz normalization.
Orthogonalizing the update reduces the cross-talk of the Hebbian write, and in the nonlinear memory successive updates are state dependent (Appendix A).

The key is that this refinement remains \emph{partial}. The incoming state already contains information accumulated from earlier chunks. Under the stock rule, each update is followed by row-wise renormalization, and repeated updates progressively rotate the state toward the current write, erasing what earlier chunks stored. DT therefore limits the number of inner updates and carries the resulting state \(M_m=M_m^{(S)}\) directly to the next chunk. The carried state is diluted, and not replaced. Multi-step partial updates thus increase \emph{capacity} while preserving one evolving memory across chunks. Yet capacity alone cannot change the structure of the write: \(k_i\) is still associated only with \(v_i\) from the same position. A relationship between different positions must therefore enter the memory in a different way.

\vspace{-0.1cm}
\subsection{Local Mixing Creates Cross-Position Bindings}
\label{subsec:DT-binding}
\vspace{-0.1cm}
Increasing capacity lets more associations survive, but it does not change which associations the memory can represent. In the standard write, \(k_i\) is paired only with \(v_i\), so the memory contains terms of the form \(v_i k_i^\top\). A relationship between two different positions is therefore absent from the write itself. No number of additional inner updates can recover a term that was never presented to the memory.

DT introduces that relationship before writing. We apply a learned depthwise causal convolution of width \(c=3\) to the query, key, and value streams
~\citep{poli2023hyena}. For the key stream,
\begin{equation}
\widetilde{k}_i
=
\sum_{r=0}^{c-1}\alpha_r^{(k)}k_{i-r},
\qquad
\sum_i v_i\widetilde{k}_i^\top
=
\sum_i\sum_{r=0}^{c-1}
\alpha_r^{(k)}v_i k_{i-r}^\top .
\label{eq:cross-write}
\end{equation}
The terms with \(r>0\) now explicitly pair information from different
positions. The same mixing is applied to the query and value streams. Thus, rather than asking the memory to infer a missing relationship after the write, DT makes that relationship part of what is written.

Importantly, the width-\(3\) convolution is not the source of long-range reach. It only determines which nearby positions become bound together. Once that relationship enters the persistent fast-weight state, it is carried across subsequent chunks and remains available to distant queries. Local mixing therefore provides the \emph{binding}; the persistent memory provides the \emph{transport}.
The memory can now store more associations and carry relationships across distance. But making a relationship available does not make it important: the memory still treats ordinary content and meaningful structural change
alike.
\vspace{-0.1cm}
\subsection{Change-Point Writes Add Selectivity}
\label{subsec:DT-selectivity}
\vspace{-0.1cm}
Binding allows relationships between positions to enter memory, but it does not determine which of them should be emphasized. As more context is written, DT therefore looks for a simple signal of change: whether the context on one side of a position differs from the context on the other. Similar surroundings should contribute little, while a strong difference should remain visible.

To expose this difference, DT views each position from both directions. The forward stream summarizes the \(w=512\) positions preceding \(i\), giving \(\mu_i^L\), while the same causal operation on the reversed sequence gives the corresponding right-context summary \(\mu_i^R\). After realigning the reverse stream, DT writes their difference,
\[
    y_i = \mu_i^L-\mu_i^R .
\]
This directly produces the desired behavior: when the two sides are similar, their summaries largely cancel; when they differ, the contrast remains. A transition from representation \(a\) to \(b\), for example, produces approximately \(a-b\), whereas a locally similar region produces a value near zero. The write therefore emphasizes change rather than treating every position equally.

The same distinction is preserved when the memory is read. If \(o_i^L\) and \(o_i^R\) are the aligned forward and reverse reads, DT retains both their shared component, \((o_i^L+o_i^R)/2\), and their directional difference, \(o_i^L-o_i^R\). The forward and reversed sequences are processed together by stacking them along the batch dimension, so the two views do not require a second sequential traversal.

The resulting memory remains causal within each directional pass. For chunk \(m\), DT first reads the state accumulated from earlier chunks,
\(O_m=f_{M_{m-1}}(\widetilde{Q}_m)\), and only then writes the current chunk to produce \(M_m\). Thus, the current write becomes available only to subsequent chunks. During masked-language pretraining, masking is applied before the query, key, and value streams are constructed, preventing a masked position from writing its own ground-truth nucleotide into memory.

    \vspace{-0.3cm}
\section{Experiments}
\label{subsec:exp-setup}
\vspace{-0.2cm}

We evaluate DT at two levels. First, controlled associative-recall experiments isolate the fast-weight memory itself, allowing the capacity and binding mechanisms to be tested independently of genomic pretraining. We then evaluate the complete model on long-context genomic tasks, where the encoder is frozen and only lightweight downstream probes are trained. This second stage tests whether the repaired memory remains useful as context grows beyond the training length, whether its write becomes selective, and whether the learned representation transfers outside the human training distribution.

Our encoder contains \(87\)M parameters in twelve identical blocks, each
combining a depthwise convolution, sliding-window attention, DT fast-weight memory, and a position-wise MLP. The embedding width is \(768\), the memory-head dimension is \(32\), and memory updates operate on chunks of \(2048\) tokens. We pretrain at single-nucleotide resolution on GRCh38 using a bidirectional masked-language-modeling objective with \(32\,\mathrm{kb}\) sequences; chromosomes 8 and 9 are excluded from pretraining and reserved for downstream human evaluation.

The sliding-window attention path covers \(128\) positions on either side per layer, giving an attention-only receptive field of approximately
\(\pm1.5\,\mathrm{kb}\) across the twelve-block stack. Information used from substantially beyond this range must therefore pass through the fast-weight memory path. Unless stated otherwise, matched comparisons use the same architecture, examples, splits, and downstream protocol; public long-context models are treated as external reference points rather than causal controls. Full dataset construction, memory variants, readout definitions, and hyperparameters are provided in Appendix~\ref{app:experiments}.

\paragraph{Readout gain.} The memory branch output is scaled by a fixed inference-time multiplier before the residual addition. This gain is not a trained parameter and does not affect pretraining; it only rescales the memory contribution at read time. We select its value on a held-in validation chromosome and hold it fixed for all reported evaluations, and we report both the amplified ($\text{gain}=4$) and unamplified ($\text{gain}=1$) traces in the context scaling experiment.

\begin{figure*}[t]
    \centering
\includegraphics[width=0.98\textwidth]{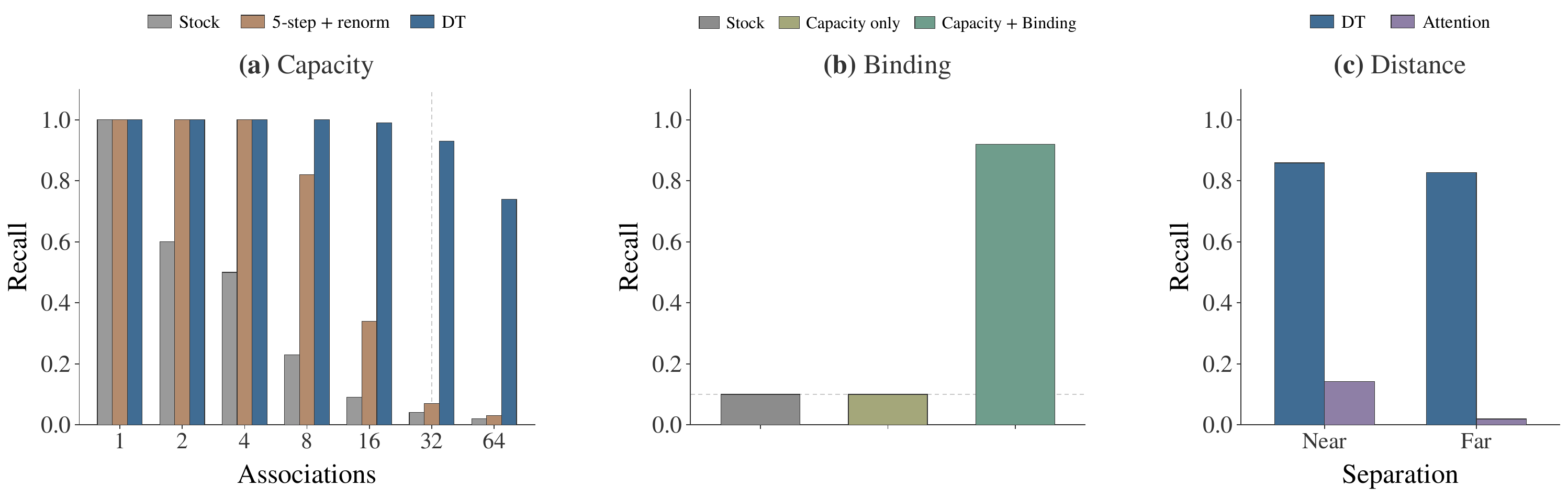}
\caption{\textbf{Capacity and binding are separate requirements for usable
memory.} \textbf{a}, Increasing associative load exposes the capacity limit of the
stock and renormalized writes, while DT remains effective up to the
memory-head dimension.
\textbf{b}, Greater capacity alone does not enable cross-position retrieval;
adding binding raises recall from chance to \(0.92\).
\textbf{c}, Once formed, the relationship remains available over distance
through the persistent memory, while the local attention path degrades
sharply.}
    \label{fig:capacity-binding}
    \vspace{-0.5cm}
\end{figure*}

\vspace{-0.1cm}
\subsection{Capacity and Binding Enable Long-Range Recall}
\label{subsec:capacity-binding}
\vspace{-0.1cm}
We first isolate DT's memory mechanism using controlled associative recall. We begin with \emph{capacity}: how many key-value associations can a \(d=32\) memory head retain as the number of stored pairs increases?
\textbf{Stock} is the original single-step large-chunk memory and provides the unrepaired baseline. \textbf{5-step + renorm} takes multiple updates while keeping the stock retention rule, separating the benefit of additional optimization from DT's update itself. \textbf{DT} uses the partial multi-step update of Sec.~\ref{subsec:DT-capacity}. Capacity, however, only tells us whether more associations
survive; it does not tell us whether information at different positions can be related. We therefore place the associated key and value at different positions and progressively compare \textbf{Stock}, \textbf{Capacity only}, and \textbf{Capacity + Binding}, where the final variant adds the cross-position mixing of Sec.~\ref{subsec:DT-binding}. Finally, successful binding could still be only local. We therefore increase the separation between the associated positions and compare DT with sliding-window attention, testing whether the relationship created locally remains available through the persistent memory.

The results follow the same progression. In Fig.~\ref{fig:capacity-binding}a, Stock degrades rapidly as associative load increases, and 5-step + renorm also collapses at higher loads, whereas DT remains near perfect through \(16\) associations and retains \(0.93\) recall at \(32\), matching the head dimension. Greater capacity alone, however, does not solve cross-position retrieval. In Fig.~\ref{fig:capacity-binding}b, both Stock and Capacity only remain at the \(0.10\) chance level, while adding binding raises recall to \(0.92\). Once that relationship is formed, it also remains available over distance. In Fig.~\ref{fig:capacity-binding}c, DT changes only from \(0.859\)
to \(0.827\) from near to far separation, whereas Attention falls from
\(0.142\) to \(0.019\). Together, these results show that usable long-range memory requires both capacity and binding: the partial multi-step update determines how much information can survive, cross-position mixing determines whether the required relationship can enter memory, and the persistent state carries that relationship beyond the local receptive field.

\begin{figure*}[t]
    \centering
    \includegraphics[width=\textwidth]
    {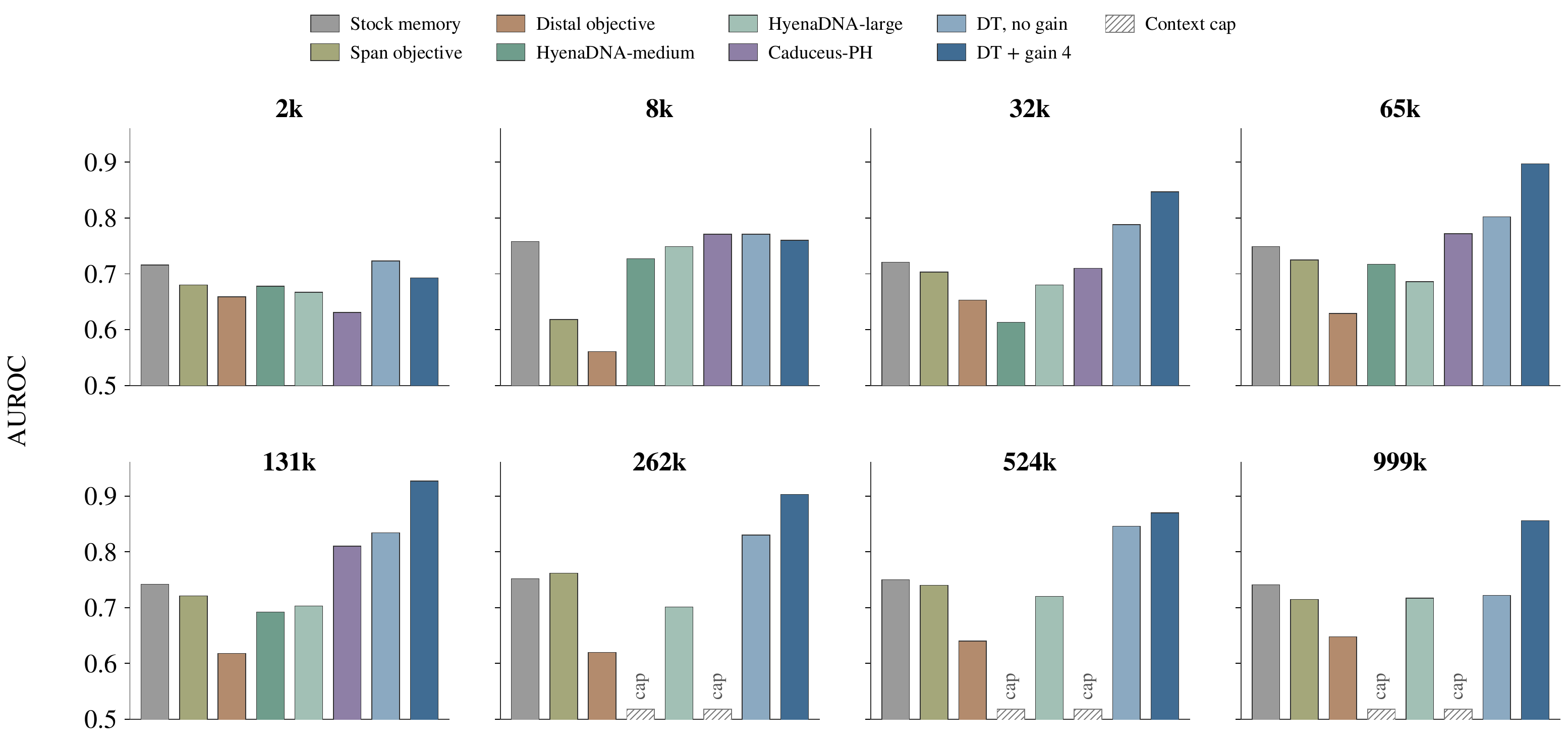}
\caption{\textbf{TAD-boundary context scaling.}
AUROC of frozen encoder features under a shared logistic-probe protocol from
\(2\,\mathrm{kb}\) to approximately \(1\,\mathrm{Mb}\). Matched controls use
the same encoder architecture, while HyenaDNA and Caduceus serve as external
long-context references. ``cap'' denotes contexts beyond a model's native
limit. Values are means over three seeds (\(0.014<\mathrm{std}<0.043\)).}
    \label{fig:tad}
    \vspace{-0.4cm}
\end{figure*}

\vspace{-0.1cm}
\subsection{DT Uses Context Beyond Its Training Length}
\label{subsec:exp-range}
\vspace{-0.1cm}
The controlled experiments show that repairing capacity and binding makes
information usable inside the fast-weight memory. We next test whether this advantage survives in a real genomic setting and, more importantly, whether the model can continue to benefit from context beyond the \(32\,\mathrm{kb}\) length seen during pretraining. We use TAD-boundary discrimination and increase the available sequence from
\(2\,\mathrm{kb}\) to approximately \(1\,\mathrm{Mb}\), while freezing every
encoder and training the same logistic probe at each context length.

We organize the eight approaches as a progressively richer sequence of ways to use long context. \textbf{Stock memory} provides the closest reference point, using the same \(87\)M-parameter encoder and downstream protocol with the original large-chunk memory. Keeping this architecture fixed, \textbf{span} masking removes contiguous regions during pretraining, while the \textbf{distal} objective removes local evidence around prediction targets, placing progressively greater pressure on the model to use information beyond its immediate neighborhood. We then move to architectures designed natively for longer genomic sequences. \textbf{HyenaDNA-medium} replaces attention with long convolutional sequence mixing and supports \(160\,\mathrm{kb}\) context;
\textbf{HyenaDNA-large} extends the same architectural principle to a native \(1\,\mathrm{Mb}\) context. \textbf{Caduceus-PH} provides a complementary bidirectional state-space architecture, allowing information from both sequence directions within a \(131\,\mathrm{kb}\) context. Finally, \textbf{DT, no gain} keeps our encoder fixed but replaces the stock memory with the repaired persistent fast-weight memory, and \textbf{DT + gain 4} adds the fixed \(4\times\) memory-branch gain used in our strongest inference setting. Contexts beyond a model's native maximum length are marked as capped in
Fig.~\ref{fig:tad}.

The results follow the same progression (Fig.~\ref{fig:tad}). With the original \textbf{Stock memory}, performance changes little as additional sequence is exposed, remaining around \(0.72\)--\(0.75\) across most context lengths. Changing only the training signal does not substantially alter this behavior: the \textbf{distal} and \textbf{span} objectives remain comparatively flat as context grows. Among the public long-context encoders, Caduceus-PH improves with context but is limited to its \(131 \, \mathrm{kb}\) native window, HyenaDNA-medium is similarly capped, and HyenaDNA-large, although it extends to \(1 \, \mathrm{Mb}\), remains relatively flat. Replacing the stock memory with DT changes the scaling pattern. \textbf{DT, no gain} improves steadily beyond the \(32\,\mathrm{kb}\) training length  through \(524 \, \mathrm{kb}\), and \textbf{DT + gain 4} strengthens the same trend, outperforming all other approaches from  \(32\,\mathrm{kb}\) onwards, peaking near \(131\,\mathrm{kb}\), and declining only gradually toward \(1\,\mathrm{Mb}\) (exact values in Table~\ref{app:context-sweep}). Thus, DT continues to benefit from sequence far beyond the context length seen during training. These results show that distant context becomes useful, but do not identify the path through which that information reaches the representation. We therefore test whether the additional signal is actually recruited through the fast-weight memory.

\vspace{-0.1cm}
\subsection{DT Retrieves Information Beyond the Local Receptive Field}
\label{subsec:recruit-distal}
\vspace{-0.1cm}

The context-scaling experiment shows that additional sequence becomes useful,
but does not by itself establish how that information reaches the
representation. We therefore use distal masking to isolate the memory path
directly. Every prediction target is separated from the nearest visible token
by at least \(3{,}072\) bp, well beyond the approximately
\(\pm1.5\,\mathrm{kb}\) receptive field of the local convolution and
sliding-window attention path. Predictive information from visible sequence
beyond this range must therefore be mediated by the persistent fast-weight
state.

With the adapting bidirectional memory enabled, DT reaches a distal validation loss of $1.292$. Freezing the same fast-weight state increases the loss to $15.30$, showing that the test-time write carries predictive information unavailable to the local path. Replacing the two directed memory passes with a single global-state control reaches $1.326$; the bidirectional formulation performs better in all six paired validation evaluations. Because the evaluation pipeline is optimized around the adapting state, we treat the frozen-state comparison as a mechanistic control rather than interpreting the absolute magnitude of the loss gap.

\vspace{-0.1cm}
\subsection{DT Selectively Preserves Genomic Change}
\label{subsec:exp-preserve}
\vspace{-0.1cm}
Having established that distant information is both useful and directly
recruited through the fast-weight state, we next ask what information that
memory preferentially preserves. We begin with
\emph{eQTL variant-effect prediction}, where the representation must capture
how a sequence change alters regulatory state. We organize the comparison from task-specific prediction to increasingly general long-context encoders.
\textbf{Task-specific expert} and \textbf{CNN} models provide supervised
references specialized for the prediction task. \textbf{HyenaDNA} introduces
long-range convolutional sequence mixing, while \textbf{Caduceus-PH} and
\textbf{Caduceus-PS} add bidirectional state-space processing over long genomic context. We then evaluate \textbf{DT} with the encoder frozen and train only a lightweight logistic adapter, asking whether the pretrained memory already exposes the information required for variant-effect prediction. Because the published baselines use their own fine-tuning protocols, we additionally use a protocol-matched frozen \textbf{HyenaDNA} comparison on whole blood. Finally, to separate sensitivity to sequence change from generic long-range decodability, we pair eQTL prediction with
\textbf{enhancer-target-gene assignment}, where the representation must retain the identity of a particular distant element rather than the effect of a local change.

\begin{figure*}[t]
    \centering
    \includegraphics[width=\textwidth]
    {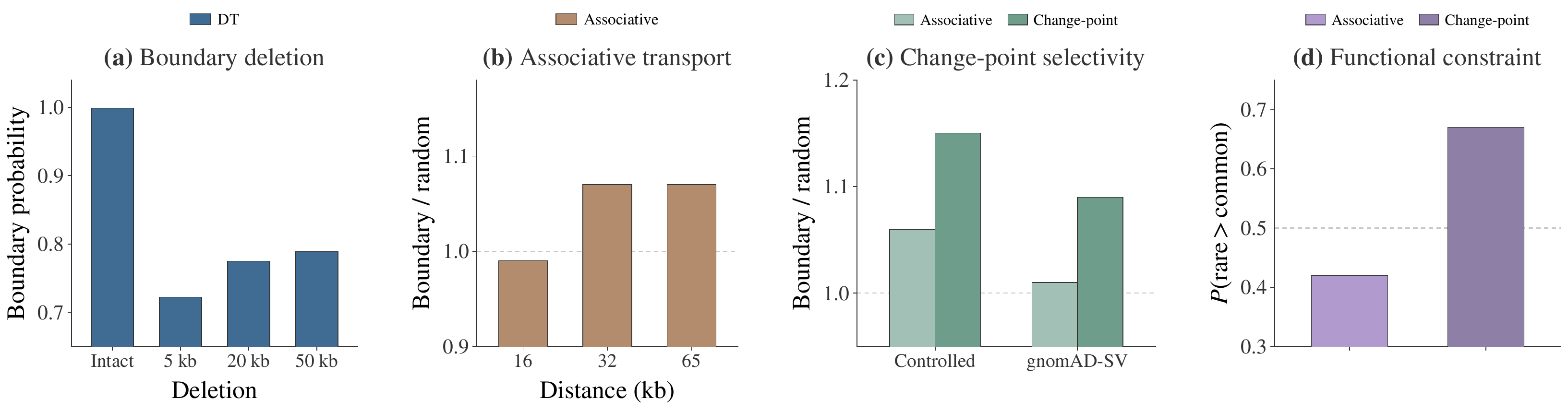}
\caption{\textbf{Selective transport of structural contrast.}
\textbf{a}, Boundary deletion changes the readout.
\textbf{b}, Associative transport is not selective.
\textbf{c}, The change-point write increases selectivity on controlled and real gnomAD variants.
\textbf{d}, It also preferentially propagates rare constrained deletions.}
    \label{fig:sv}
   \vspace{-0.4cm} 
\end{figure*}

We then move from downstream decoding to increasingly direct tests of what the memory carries. At \(262\,\mathrm{kb}\) context, we first delete
\(5\), \(20\), and \(50\,\mathrm{kb}\) regions centered on annotated TAD
boundaries and measure the resulting change in the boundary readout
(Fig.~\ref{fig:sv}a). We next hold the distal readout fixed and compare
boundary-centered perturbations with matched random perturbations over
\(16\)-\(65\,\mathrm{kb}\). Using the standard \textbf{associative write}
first establishes how the repaired capacity-and-binding memory transports these distant perturbations (Fig.~\ref{fig:sv}b). We then replace only the write target with the \textbf{bidirectional change-point write} and repeat the comparison, first on controlled boundaries and then on real gnomAD structural variants (Fig.~\ref{fig:sv}c). Finally, we compare rare constrained deletions with common tolerated variants, extending the same comparison from structural contrast to functional constraint (Fig.~\ref{fig:sv}d). The nine GTEx tissues are cultured fibroblasts (CCF), whole blood (WB), thyroid (Thy), non-sun-exposed skin (SNSE), sun-exposed lower-leg skin (SSEL), skeletal muscle (Musc), tibial nerve (Nerv), tibial artery (Art), and subcutaneous adipose tissue (Adip).

\begin{table*}[t]
    \centering
    \caption{\textbf{eQTL variant-effect prediction across nine GTEx tissues.}
    AUROC across tissue-specific tasks. DT uses a frozen encoder with a logistic
    adapter; published baselines use their reported fine-tuning protocols.
    Bold denotes the best result in each column.}
    \label{tab:eqtl-tissues}
    \small
    \setlength{\tabcolsep}{4pt}
    \begin{tabular}{lccccccccc c}
        \toprule
        \textbf{Model} & CCF & WB & Thy & SNSE & SSEL & Musc & Nerv & Art & Adip & \textbf{Avg} \\
        \midrule
        Task-specific expert
        & $.639$ & $.689$ & $.612$ & $.710$ & $.700$ & $.621$ & $.683$
        & $\mathbf{.741}$ & $\mathbf{.736}$ & $.681$ \\

        CNN
        & $.547$ & $.577$ & $.487$ & $.499$ & $.499$ & $.502$ & $.516$
        & $.576$ & $.551$ & $.528$ \\

        HyenaDNA (fine-tuned)
        & $.584$ & $.512$ & $.529$ & $.471$ & $.544$ & $.487$ & $.511$
        & $.479$ & $.513$ & $.514$ \\

        Caduceus-Ph (fine-tuned)
        & $.597$ & $.594$ & $.527$ & $.586$ & $.574$ & $.538$ & $.588$
        & $.547$ & $.541$ & $.565$ \\

        Caduceus-PS (fine-tuned)
        & $.549$ & $.542$ & $.547$ & $.529$ & $.541$ & $.523$ & $.552$
        & $.536$ & $.519$ & $.537$ \\

        \midrule

        \textbf{DT (ours, frozen)}
        & $\mathbf{.745}$ & $\mathbf{.747}$ & $\mathbf{.638}$
        & $\mathbf{.780}$ & $\mathbf{.829}$ & $\mathbf{.729}$
        & $\mathbf{.763}$ & $.682$ & $.640$ & $\mathbf{.728}$ \\

        \bottomrule
    \end{tabular}
    \vspace{-0.4cm}
\end{table*}

The results follow this progression. Frozen DT reaches a mean AUROC of
\(0.728\) across nine GTEx tissues and \(0.747\) in the matched whole-blood
comparison versus \(0.689\) for frozen HyenaDNA; on enhancer--target-gene
assignment, the ordering reverses (\(0.628\) versus \(0.724\)). In Fig.~\ref{fig:sv}a, deleting the centered
\(5\,\mathrm{kb}\) boundary region produces the largest drop in boundary
probability (\(0.277\)), compared with \(0.224\) and \(0.210\) for the
\(20\) and \(50\,\mathrm{kb}\) deletions. In Fig.~\ref{fig:sv}b, the
associative write remains close to no preference across distance
(\(0.99, 1.07, 1.07\)). Adding the change-point write increases the
boundary-to-random ratio from \(1.06\) to \(1.15\) on controlled boundaries
and from \(1.01\) to \(1.09\) on gnomAD-SV (Fig.~\ref{fig:sv}c).
Finally, Fig.~\ref{fig:sv}d shows the strongest separation:
\(P(\mathrm{rare}>\mathrm{common})\) increases from \(0.42\) with the
associative write to \(0.67\) with the change-point write.

\vspace{-0.1cm}
\subsection{DT Transfers Beyond Human Genomes}
\label{subsec:mouse}
\vspace{-0.1cm}

\begin{table}[t]
\centering
\small
\caption{\textbf{Zero-shot cross-species transfer.}
TAD-boundary AUROC from frozen human-pretrained encoders under a matched
logistic-probe protocol. Human results use held-out chromosomes 8/9; mouse
results use mm10 chromosomes 1-5 without adaptation. ``cap'' denotes context
beyond a model's native limit.}

\label{tab:crossspecies}
\setlength{\tabcolsep}{4pt}
\begin{tabular}{llccccccc}
\toprule
 & \textbf{Encoder} & \textbf{2k} & \textbf{8k} & \textbf{32k}
 & \textbf{131k} & \textbf{262k} & \textbf{524k} & \textbf{1M} \\
\midrule

\multirow{4}{*}{\rotatebox{90}{Human}}
 & DT, memory on
 & \textbf{.693} & \textbf{.760} & \textbf{.847}
 & \textbf{.927} & \textbf{.903} & \textbf{.870} & \textbf{.856} \\

 & Same encoder, memory off
 & .640 & .641 & .617 & .623 & .658 & .625 & .610 \\

 & Caduceus-PS
 & .668 & .739 & .782 & .827 & cap & cap & cap \\

 & HyenaDNA-1m
 & .633 & .701 & .640 & .660 & .776 & .712 & .784 \\

\midrule

\multirow{4}{*}{\rotatebox{90}{Mouse}}
 & DT, memory on
 & .677 & .699 & \textbf{.786}
 & \textbf{.785} & \textbf{.761} & \textbf{.783} & \textbf{.796} \\

 & Same encoder, memory off
 & .653 & .583 & .602 & .639 & .630 & .648 & .713 \\

 & Caduceus-PS
 & \textbf{.710} & .697 & .752 & .773 & cap & cap & cap \\

 & HyenaDNA-1m
 & .615 & \textbf{.711} & .693 & .733 & .713 & .698 & .710 \\

\bottomrule
\end{tabular}
\vspace{-0.4cm}
\end{table}

\begin{table}[t]
\centering
\small
\caption{\textbf{Cross-species representation alignment.}
Normalized centroid distance from each species mean window embedding to the
human embedding centroid, before and after test-time adaptation. Lower is
closer; parentheses denote evolutionary distance from human in millions of years(M).}
\label{tab:species-adaptation}
\setlength{\tabcolsep}{7pt}
\begin{tabular}{lccccc}
\toprule
& \textbf{Human} & \textbf{Chimp} & \textbf{Macaque}
& \textbf{Mouse} & \textbf{Chicken} \\
& $(0)$ & $(6.4M)$ & $(29M)$ & $(87M)$ & $(319M)$ \\
\midrule

\textbf{Frozen}
& $.000$
& $.090$
& $.133$
& $.286$
& $.112$ \\

\textbf{Adapted}
& $.000$
& $.097$
& $.136$
& $.271$
& $.120$ \\

\bottomrule
\end{tabular}
\vspace{-0.4cm}
\end{table}

Having established capacity, binding, and selective retention on human genomic sequence, we finally test whether the resulting long-range representation survives distribution shift. We begin with a controlled human-to-mouse transfer setting in which the task and downstream protocol remain fixed while the underlying genome changes. All encoders are pretrained on human sequence and then frozen. Human evaluation uses held-out chromosomes 8 and 9, while mouse evaluation uses mm10 chromosomes 1--5 with no encoder adaptation. We sweep context from \(2\,\mathrm{kb}\) to \(1\,\mathrm{Mb}\) using the same logistic-probe protocol in both species. \textbf{Memory off} uses the same DT encoder with its fast-weight path disabled, providing the closest reference for how much of the transfer comes from long-range memory.
\textbf{Caduceus-PS} adds bidirectional state-space processing with a native
\(131\,\mathrm{kb}\) context, while \textbf{HyenaDNA-1m} provides a native
megabase convolutional long-context model. \textbf{DT, memory on} retains the
full persistent fast-weight memory, allowing us to compare these increasingly
long-range representations under the same human-to-mouse shift
(Table~\ref{tab:crossspecies}).

We then increase the difficulty of the transfer itself. Rather than evaluating only one target species, we move progressively farther from the human training distribution through chimpanzee, macaque, mouse, and finally chicken (Table ~\ref{tab:species-adaptation}). At each shift, we compare the
human-pretrained representation with its test-time-adapted memory state. This
second setting therefore asks not only whether the representation transfers,
but how its behavior changes as evolutionary distance from the training genome increases.

The zero-shot results preserve DT's long-context behavior under the
human-to-mouse shift. On mouse, DT reaches \(0.786\) AUROC at
\(32\,\mathrm{kb}\) and remains \(0.785\), \(0.761\), \(0.783\), and
\(0.796\) from \(131\,\mathrm{kb}\) to \(1\,\mathrm{Mb}\). Over the same
range, the memory-off encoder remains lower, while Caduceus-PS reaches
\(0.773\) at its \(131\,\mathrm{kb}\) native limit and HyenaDNA-1m reaches
\(0.733\)-\(0.710\) from \(131\,\mathrm{kb}\) to \(1\,\mathrm{Mb}\)
(Table~\ref{tab:crossspecies}). Thus, the long-context representation learned on human sequence remains usable after zero-shot transfer to mouse. Across the broader species sweep, centroid distance from human increases with evolutionary distance among mammals, while chicken departs from this trend. Frozen and adapted representations remain similar throughout (Table ~\ref{tab:species-adaptation}).



\vspace{-0.3cm}
\section{Conclusion and Limitations}
\vspace{-0.2cm}
Processing a long sequence is not sufficient to remember it. We identify \emph{capacity}, \emph{binding}, and \emph{selectivity} as three requirements for usable long-range test-time memory, and DT addresses them through multi-step updates, cross-position mixing, and bidirectional change-point
writes. Controlled experiments isolate each mechanism, while genomic results show that the resulting memory remains useful far beyond its $32$\,kb training context and transfers across species at megabase scale.

Our study is limited to genomic sequence and does not incorporate transcriptomic, epigenomic, or proteomic supervision; comparison with some recent TTT methods is also constrained by the lack of reproducible implementations. Extending these memory principles across modalities, architectures, and domains is an important next step. More broadly, long-context modeling should not only ask how much context a model can process, but how much of that context it can retain, relate, and use.

    \bibliographystyle{iclr2027_conference}
    \bibliography{iclr_2027}

    \appendix
    

\section*{Appendix}

\section{Defects In Stock TTT}

We observe two distinct defects in stock TTT, and show that our implementation of DT addresses these defects. 
\paragraph{Defect 1: a single inner step gives Hebbian capacity.}
Write the fast weights as a linear map $W$ over a chunk of $N$ key--value pairs
$K = [k_1,\dots,k_N]$, $V = [v_1,\dots,v_N]$, with unit-norm keys and the dot-product objective
\begin{equation}
  \mathcal{L}(W) = -\sum_{i=1}^{N} v_i^\top W k_i, \qquad \nabla_W \mathcal{L} = -VK^\top .
  \label{eq:dot-objective}
\end{equation}
A single step from the carried state $W_0$ gives $W_1 = W_0 + \eta VK^\top$, whose readout at a stored key is
\begin{equation}
  W_1 k_j = W_0 k_j + \underbrace{\eta\, v_j}_{\text{signal}}
  + \underbrace{\eta \sum_{i \neq j} (k_i^\top k_j)\, v_i}_{\text{cross-talk}} .
\end{equation}

For orthonormal values and keys uniform on the sphere, the expected cross-talk energy relative to the
signal is $(N-1)/d$, growing linearly with load.  Because the gradient in Eq.~\eqref{eq:dot-objective} does not depend on $W$, repeating the step $S$ times on the same chunk without normalization adds the same $VK^\top$ each time and is therefore equivalent to one step with rate $S\eta$. Additional plain steps rescale the write but leave the signal to cross-talk ratio unchanged. Newton-Schulz normalization instead replaces $VK^\top$ with its orthogonal polar factor $P = V(K^\top K)^{-1/2}K^\top$, whose readout at a stored key is:

\begin{equation}
  P k_j = V (K^\top K)^{1/2} e_j ,
\end{equation}
replacing the Hebbian coefficients $K^\top K = I + E$ with $(I+E)^{1/2} \approx I + E/2$; for weakly
correlated keys this reduces cross-talk energy about fourfold at nearly the same signal level.

Retention acts on the carried state. The stock rule renormalizes each row of $W$ after every update;
repeated steps in a fixed direction progressively rotate each row toward the new write, erasing $W_0$
in the limit. Free retention removes this renormalization, so $W_0$ is diluted rather than replaced.
For the nonlinear SwiGLU memory the gradient depends on $W$, and the benefit of multiple steps there
is empirical (Fig.~2a).

    \paragraph{Defect 2: the write is self-associative.} Capacity is necessary and not sufficient. Both $k_i$ and $v_i$ are projections of the same position $i$,
    and every write in Defect~1 takes the form $V\,C\,K^\top$, with $C = I$ for gradient descent and
    $C = (K^\top K)^{-1/2}$ for the orthogonalized update. Any cross-position coefficient is therefore fixed
    by the key Gram matrix rather than by which positions should be related: the readout at a stored key is
    $V\,g(K^\top K)\,e_i$ with $g(x) = x$ or $x^{1/2}$, and for orthonormal keys it returns exactly $v_i$.
    No inner rate or step count changes this, so a retrieval requiring cross-token binding is not
    expressible, whatever the capacity.
     
    A causal convolution of width $c$ applied to the memory's projections before the
    write removes the restriction. With
    $\tilde{k}_i = \sum_{s=0}^{c-1} \alpha_s k_{i-s}$, the outer product becomes
    \begin{equation}
    \sum_{i} v_i \tilde{k}_i^{\!\top}
    \;=\; \sum_{i}\sum_{s=0}^{c-1} \alpha_s\, v_i\, k_{i-s}^{\!\top},
    \label{eq:conv}
    \end{equation}
    which contains cross-token terms $v_i k_{i-s}^{\!\top}$ for $s \geq 1$. The
    convolution supplies the local mixing that makes cross-token binding expressible;
    at $c=1$ it is the identity and Eq.~\eqref{eq:conv} reduces to the
    self-associative case.
     
    With a multi-step unpinned inner update: five steps, free retention. recall is
    $0.99$ at sixteen associations and $0.93$ at thirty-two, against a head dimension
    of $32$ (Fig.~\ref{fig:capacity-binding}a): the memory saturates at approximately the
    head dimension, whereas stock recall already falls to $0.60$ at two associations (Table~\ref{tab:app-capacity}).

\section{Detailed Experimental Setup}
\label{app:experiments}

This appendix provides the complete experimental protocol underlying the
results in the main paper. We distinguish three components throughout:
(i) the pretrained sequence encoder, (ii) the test-time memory configuration,
and (iii) the downstream readout. Unless explicitly stated otherwise,
downstream genomic evaluations freeze the pretrained encoder and train only a
lightweight task-specific probe. Modifications to DT therefore change the
memory used during the forward pass, not the pretrained model parameters.

\subsection{Model Architecture}
\label{app:architecture}

DT is evaluated within an $87$\,M-parameter nucleotide-level encoder composed
of twelve pre-normalized blocks. Each block contains a depthwise 1D
convolution, bidirectional sliding-window attention, a fast-weight memory
module, and a position-wise MLP. The embedding width is $768$, with $24$
attention and memory heads of dimension $32$.

\begin{table}[h]
    \centering
    \caption{\textbf{Encoder architecture.}
    Architecture shared by the matched DT and stock-memory models.}
    \label{tab:app-architecture}
    \small
    \setlength{\tabcolsep}{5pt}
    \begin{tabular}{ll}
        \toprule
        \textbf{Component} & \textbf{Configuration} \\
        \midrule
        Parameters & $\approx87$\,M \\
        Blocks & $12$ pre-LN blocks \\
        Embedding width & $768$ \\
        Attention heads & $24 \times 32$ \\
        Memory heads & $24 \times 32$ \\
        Attention window & $\pm128$ nt per block \\
        Attention-only reach & $\pm1{,}536$ nt across 12 blocks \\
        Memory chunk size & $2{,}048$ tokens \\
        Position encoding & RoPE, base $10^4$ \\
        MLP & $768 \rightarrow 3072 \rightarrow 768$ \\
        MLP activation & GELU \\
        Vocabulary & A, C, G, T, N, \texttt{[MASK]} \\
        Input resolution & single nucleotide \\
        Pretraining context & $32{,}768$ tokens \\
        Pretraining genome & GRCh38 \\
        Held-out chromosomes & chr8, chr9 \\
        Objective & bidirectional masked language modeling \\
        \bottomrule
    \end{tabular}
\end{table}

The sliding-window attention path can communicate at most $128$ positions per
layer. Across twelve blocks, its maximum receptive field is therefore
$\pm1{,}536$ nucleotides. Information reaching a central readout from
substantially beyond this range must use the fast-weight memory path.

\paragraph{Encoder and output head.}
Input nucleotides are embedded into $768$ dimensions and processed by the
twelve-block encoder. During pretraining, the final LayerNorm output is mapped
through a linear $768\!\rightarrow\!6$ nucleotide-classification head. This
head is used only for masked-language modeling. Downstream experiments extract
the encoder representation \emph{before} this vocabulary projection and train
a separate task-specific probe. There is therefore no separate autoregressive
decoder in the downstream experiments.

\paragraph{Fast-weight network.}
Each memory head uses a three-matrix SwiGLU fast-weight function,
\[
f_M(x)
=
M_1\!\left[
    \operatorname{SiLU}(M_0x)\odot M_2x
\right],
\]
with $M_0,M_1,M_2\in\mathbb{R}^{32\times32}$. Queries and keys are
$\ell_2$-normalized before entering the memory. The memory state is updated
during the forward pass and subsequently read by the query stream.

\subsection{DT Memory Configurations}
\label{app:memory-variants}

The experiments progressively modify the same underlying fast-weight memory.
This design separates the three properties studied in the paper---capacity,
cross-position binding, and selective retention---without changing the
surrounding encoder.

\begin{table}[h]
    \centering
    \caption{\textbf{Memory configurations.} Matched variants share the
    encoder and differ only in the marked operations. MS: multi-step;
    FR: free retention; LM: local mixing; CP: change-point write.}
    \label{tab:app-memory}
    \footnotesize
    \setlength{\tabcolsep}{4pt}
    \begin{tabular}{lccccl}
        \toprule
        \textbf{Configuration} & \textbf{MS} & \textbf{FR} & \textbf{LM} & \textbf{CP} & \textbf{Role} \\
        \midrule
        Stock memory       &            &            &            &            & Matched control \\
        Capacity repair    & \checkmark & \checkmark &            &            & Capacity test \\
        Capacity + binding & \checkmark & \checkmark & \checkmark &            & Binding test \\
        DT, no gain        & \checkmark & \checkmark & \checkmark & & No readout gain \\
        DT + gain 4        & \checkmark & \checkmark & \checkmark &   & $4\times$ memory gain at inference \\
        Selective DT       & \checkmark & \checkmark & \checkmark & \checkmark & Selectivity experiments \\
        \bottomrule

    \end{tabular}
\end{table}

\paragraph{Partial update.}
DT starts each chunk from the memory carried by the preceding chunk and
performs five Newton--Schulz-normalized inner updates. Unlike the stock
renormalized configuration, the resulting state is not projected back to a
prescribed norm after each chunk. The finite update budget limits how far a
single chunk can move the persistent state while allowing the current
associations to be fitted more accurately.

\paragraph{Cross-position binding.}
Before the memory write, a learned depthwise causal convolution of width $3$
is applied to the query, key, and value streams. This operation is deliberately
local: it does not itself provide long-range communication. Instead, it makes
cross-position relationships available to the write; the persistent fast-weight
state subsequently transports those relationships across chunks.

\subsection{Pretraining Data and Objective}
\label{app:pretraining}

The model operates at single-nucleotide resolution with the vocabulary
$\{\mathrm{A,C,G,T,N,[MASK]}\}$. Any unsupported nucleotide symbol is mapped
to N. Pretraining uses GRCh38 sequences of length $32{,}768$ nucleotides, with
chromosomes 8 and 9 excluded from training and reserved for downstream human
evaluation.

The primary model is pretrained with bidirectional masked-language modeling.
Two matched stock-memory controls additionally use long-range masking
objectives:

\begin{description}
    \item[Token MLM.]
    Individual non-N positions are selected with probability $0.15$ and use
    standard masked-token corruption.

    \item[Span MLM.]
    Contiguous spans are masked until approximately $15\%$ of the sequence is
    selected.

    \item[Distal MLM.]
    Prediction targets are selected while masking their local neighborhood,
    forcing the prediction to depend on information outside the
    attention-only receptive field.
\end{description}

The span and distal models are used only as objective controls. They test
whether adding explicitly long-range pretraining supervision is sufficient to
recover the long-context behavior of DT without repairing the memory
mechanism.

\subsection{Comparison Models}
\label{app:baselines}

We separate \emph{matched controls}, which retain the \(87\)M DT architecture,
from \emph{external baselines}, which provide long-context genomic reference
points but differ in architecture, scale, and pretraining.

\begin{table*}[h]
    \centering
    \caption{\textbf{Comparison models.}
    Matched controls preserve the DT architecture; public models serve as
    external long-context references.}
    \label{tab:app-baselines}
    \small
    \setlength{\tabcolsep}{5pt}
    \begin{tabular}{llllp{4.6cm}}
        \toprule
        \textbf{Model}
        & \textbf{Params.}
        & \textbf{Context}
        & \textbf{Protocol}
        & \textbf{Purpose} \\
        \midrule

        DT
        & \(87\)M
        & \(32\) kb
        & frozen
        & Full repaired fast-weight memory. \\

        Stock memory
        & \(87\)M
        & \(32\) kb
        & frozen
        & Architecture-matched memory control. \\

        Stock + distal
        & \(87\)M
        & \(32\) kb
        & frozen
        & Tests distal supervision alone. \\

        Stock + span
        & \(87\)M
        & \(32\) kb
        & frozen
        & Tests span masking alone. \\

        HyenaDNA-medium
        & \(\sim7\)M
        & \(160\) kb
        & public
        & Long-context convolutional reference. \\

        HyenaDNA-large
        & \(\sim7\)M
        & \(1\) Mb
        & public
        & Megabase-context reference. \\

        Caduceus-PH
        & \(\sim8\)M
        & \(131\) kb
        & public
        & Bidirectional state-space reference. \\

        Caduceus-PS
        & \(\sim8\)M
        & \(131\) kb
        & public
        & Cross-species state-space reference. \\

        \bottomrule
    \end{tabular}
\end{table*}

Because the public models are substantially smaller than DT, we do not use
their comparison to establish the causal effect of the proposed memory
modification. That conclusion instead rests on the matched stock-memory
ablation. External models establish the scale of performance achieved by
public long-context genomic encoders.

\subsection{Downstream Readouts}
\label{app:readouts}

Unless otherwise stated, genomic evaluations freeze all pretrained encoder
parameters. Task performance is measured using lightweight readouts trained on
the resulting representations.

\begin{table*}[h]
    \centering
    \caption{\textbf{Evaluation interfaces.}
    The encoder remains frozen in the genomic downstream evaluations.}
    \label{tab:app-readouts}
    \small
    \begin{tabular}{llll}
        \toprule
        \textbf{Experiment}
        & \textbf{Encoder}
        & \textbf{Readout}
        & \textbf{Metric} \\
        \midrule

        Associative capacity
        & isolated memory
        & direct retrieval
        & recall \\

        Cross-position binding
        & isolated memory
        & direct retrieval
        & recall \\

        TAD boundary
        & frozen
        & pooled logistic probe
        & AUROC \\

        eQTL variant effect
        & frozen
        & logistic / linear adapter
        & AUROC \\

        Enhancer--target gene
        & frozen
        & matched frozen adapter
        & AUROC \\

        Structural perturbation
        & frozen
        & boundary readout
        & probability change \\

        Selective transport
        & frozen
        & distal masked readout
        & boundary/random ratio \\

        Human$\rightarrow$mouse TAD
        & frozen human encoder
        & logistic probe
        & AUROC \\

        \bottomrule
    \end{tabular}
\end{table*}

\subsection{Controlled Memory Experiments}
\label{app:synthetic}

The controlled experiments isolate the fast-weight memory from genomic
pretraining and test the two mechanism-level failures identified in the main
paper.

\paragraph{Associative capacity.}
We construct multi-query associative-recall instances containing
$N\in\{1,2,4,8,16,32,64\}$ key--value associations. The memory-head dimension
is $d=32$. Each method receives the same associations and queries; recall is
measured by recovering the corresponding stored value. We compare the stock
single-step write, a five-step write with state renormalization, and DT's
partial multi-step update. This experiment tests storage capacity independently
of cross-position binding.

\paragraph{Cross-position binding.}
We next construct an end-to-end associative task in which the required key and
value occur at different positions. The task has ten candidate outputs, giving
chance recall of $0.10$. The stock and capacity-only configurations test whether
additional storage alone is sufficient; adding width-$3$ mixing tests whether
the missing cross-position relationship must first be made representable to the
memory.

\paragraph{Distance.}
Finally, we increase the separation between the associated positions while
keeping the local mixing operator unchanged. We compare DT with the
sliding-window attention path. Because the convolution itself is local,
persistence at the far separation tests whether the resulting association is
being transported by the fast-weight state rather than by the mixing operator.

\subsection{TAD-Boundary Context Scaling}
\label{app:tad}

\paragraph{Annotations.}
Human TAD boundaries are derived from the GM12878 Arrowhead domain calls of
Rao et al.\ (2014). Domain coordinates are lifted from hg19 to hg38; a domain
is retained only when both endpoints map successfully to the same target
chromosome and its lifted length remains within $20\%$ of the original.
Shared endpoints between adjacent domains are collapsed into a single
boundary.

\paragraph{Evaluation chromosomes.}
Human evaluation uses chromosomes 8 and 9, which are held out from model
pretraining. All encoders receive the same sequence windows and downstream
labels for a given evaluation condition.

\paragraph{Window construction.}
For a context length $L$, positive examples are centered on annotated TAD
boundaries. Negative controls are centered at positions at least
$100$\,kb from the nearest annotated boundary. Windows are required to fit
inside the source sequence and contain at most $20\%$ N bases. Classes are
balanced.

\paragraph{Feature extraction.}
The frozen encoder produces hidden states in
$\mathbb{R}^{L\times768}$. We extract the central $512$ positions and compute
both mean and max pooling over the sequence axis, yielding a
$1{,}536$-dimensional representation. Public baselines use the same pooling
operation on their final hidden states.

\paragraph{Logistic probe.}
Features are standardized using statistics from the training split. A single
linear logit is optimized with binary cross-entropy. The probe is trained for
400 full-batch Adam steps with learning rate $0.1$ and weight decay
$10^{-3}$. Evaluation uses AUROC.

\paragraph{Context sweep.}
We evaluate contexts from $2$\,kb to approximately $1$\,Mb. The pretrained
DT parameters are fixed throughout: context length is changed only at
inference. Consequently, all cells beyond $32$\,kb measure extrapolation beyond
the context length used during pretraining. 
\paragraph{Interpretation.}
This task measures discrimination between boundary-centered windows and
windows sampled far from any boundary. It should therefore not be interpreted
as calibrated genome-wide boundary calling. In particular, positions at
intermediate distances from an annotated boundary are not represented by
either training extreme. This limitation affects the absolute interpretation
of the AUROC but not the comparison between encoders evaluated on identical
examples.

\subsection{DNALongBench Variant-Effect Evaluation}
\label{app:dnalongbench}

We use paired downstream tasks to test whether the frozen representation favors
structural \emph{contrast} over precise long-range \emph{identity}. All
encoders use the same examples and downstream adapter within the
protocol-matched comparison.

\paragraph{eQTL variant effect.}
The task asks whether replacing an allele changes regulatory state. DT is
evaluated as a frozen encoder using a lightweight logistic adapter. Whole blood
also provides a protocol-matched comparison against frozen HyenaDNA. We
additionally evaluate the same frozen DT representation across nine GTEx
tissues: cultured fibroblasts, whole blood, thyroid, non-sun-exposed skin,
sun-exposed lower-leg skin, skeletal muscle, tibial nerve, tibial artery, and
subcutaneous adipose.

The task-specific expert, CNN, fine-tuned HyenaDNA, and fine-tuned Caduceus
numbers in Table~\ref{tab:eqtl-tissues} are published DNALongBench references.
They use their respective published training protocols and are therefore
reported to establish scale rather than as protocol-matched encoder
comparisons.

\paragraph{Enhancer--target-gene assignment.}
This task asks which distant enhancer is associated with a specific gene and
therefore requires preserving long-range element identity. We use the same
frozen-encoder evaluation principle as above. The contrast between this task
and eQTL effect prediction is used to test the contrast-versus-identity
hypothesis in Sec.~\ref{subsec:exp-preserve}.

\subsection{Structural Perturbation Experiments}
\label{app:structural}

We next test whether the long-range representation responds specifically to
boundary-scale genomic structure rather than to arbitrary sequence removal.

\paragraph{Boundary deletion.}
For $100$ annotated boundaries at $262$\,kb context, we compare the intact
sequence with variants in which $5$, $20$, or $50$\,kb centered regions are
deleted. The intact windows have predicted boundary probability
$0.999\pm0.001$. We report the change relative to the corresponding intact
sequence, yielding drops of $0.277$, $0.224$, and $0.210$ for the three
deletion sizes.

\subsection{Selective Long-Range Transport}
\label{app:selectivity}

Boundary recruitment does not by itself imply selective transport. We therefore
measure whether perturbing a functional boundary affects a distant readout more
strongly than perturbing a matched random region.

For each sequence, the distal readout location is held fixed while the
perturbation is applied outside its local attention receptive field. We compute
\[
    S_{\mathrm{sel}}
    =
    \frac{\Delta_{\mathrm{boundary}}}
         {\Delta_{\mathrm{random}}},
\]
where $S_{\mathrm{sel}}>1$ indicates preferential transport of the boundary
perturbation.

\paragraph{Associative write.}
With the standard associative target, the ratio remains close to one:
$0.99$, $1.07$, and $1.07$ at $16$, $32$, and $65$\,kb, respectively.
Thus the repaired associative memory can transport distant changes without
preferentially preserving boundary structure.

\paragraph{Functional constraint.}
Finally, we compare constrained rare deletions with common tolerated variants.
The probability that the rare variant produces the larger propagated effect is
$0.67$ for the change-point write and $0.42$ for the associative write. This
experiment tests whether the selectivity introduced by the change-point target
tracks functional constraint rather than perturbation magnitude alone.


\section{Extended Experimental Details}
\label{app:experiments}

This section provides the experiment-specific protocols underlying
Sec.~\ref{subsec:exp-range}. Shared architecture, pretraining, memory
configuration, and downstream-readout details are given above; here we focus
on how each experiment is constructed, what is held fixed, and what each
comparison measures. We follow the same progression as the main paper:
controlled capacity and binding, long-context extrapolation, selective
transport, and finally cross-species transfer.

\subsection{Capacity and Binding Experiments}
\label{app:capacity-binding}

The controlled experiments isolate the fast-weight memory from genomic
pretraining. Their purpose is to separate three questions that are otherwise
coupled in a full sequence model: how many associations can be retained,
whether information at different positions can be bound, and whether an
association created locally remains accessible over distance.

\paragraph{Associative capacity.}
We construct multi-query associative-recall instances containing
\[
N\in\{1,2,4,8,16,32,64\}
\]
key--value associations. The memory-head dimension is fixed at \(d=32\).
Every method receives the same associations and queries, and performance is
measured by recall of the corresponding stored value.

We compare three increasingly permissive writes. \textbf{Stock} uses the
original single-step large-chunk update. \textbf{5-step + renorm} performs
five updates but retains the stock state-renormalization rule, separating the
effect of repeated optimization from persistent retention.
\textbf{DT} uses five partial updates with free retention, carrying the
resulting fast-weight state forward without projecting it back to the stock
prescribed norm.

\begin{table}[t]
\centering
\small
\caption{\textbf{Associative capacity.}
Recall as the number of simultaneously stored key--value pairs increases.
The memory-head dimension is \(d=32\).}
\label{tab:app-capacity}
\setlength{\tabcolsep}{5.5pt}
\begin{tabular}{lccccccc}
\toprule
\textbf{Method}
& \(\mathbf{1}\) & \(\mathbf{2}\) & \(\mathbf{4}\)
& \(\mathbf{8}\) & \(\mathbf{16}\) & \(\mathbf{32}\)
& \(\mathbf{64}\) \\
\midrule
Stock
& 1.00 & .60 & .50 & .23 & .09 & .04 & .02 \\
5-step + renorm
& 1.00 & 1.00 & 1.00 & .82 & .34 & .07 & .03 \\
\textbf{DT}
& 1.00 & 1.00 & 1.00 & 1.00 & .99 & \textbf{.93} & \textbf{.74} \\
\bottomrule
\end{tabular}
\end{table}

The comparison separates repeated fitting from retention. Repeated updates
substantially improve the low-load regime, but the renormalized state again
degrades as associative load approaches the head dimension. DT remains near
perfect through \(N=16\) and retains \(0.93\) recall at \(N=32\).

\paragraph{Cross-position binding.}
Capacity alone does not test whether the memory can represent an association
between different sequence positions. We therefore construct an end-to-end
retrieval task in which the required key and value occur at different
positions. Each example contains ten possible outputs, giving chance recall
of \(0.10\).

We compare \textbf{Stock}, \textbf{Capacity only}, and
\textbf{Capacity + Binding}. The first two differ in storage capacity but
retain the same-position write. The final configuration additionally applies
the width-\(3\) cross-position mixing described in
Sec.~\ref{subsec:DT-binding}.

\begin{table}[t]
\centering
\small
\caption{\textbf{Cross-position binding.}
End-to-end recall when the associated key and value occur at different
positions. Chance performance is \(0.10\).}
\label{tab:app-binding}
\begin{tabular}{lc}
\toprule
\textbf{Configuration} & \textbf{Recall} \\
\midrule
Stock & .10 \\
Capacity only & .10 \\
Capacity + Binding & \textbf{.92} \\
\bottomrule
\end{tabular}
\end{table}

The capacity repair alone therefore leaves the cross-position task at chance,
whereas introducing the relationship before the write raises recall to
\(0.92\).

\paragraph{Distance.}
The binding operator itself is deliberately local. We therefore increase the
separation between the associated positions while leaving the width-\(3\)
mixing operator unchanged. DT is compared with the sliding-window attention
path under the same retrieval task. Persistence at the farther separation
therefore tests whether an association created locally is subsequently carried
by the persistent memory rather than by the local mixing operation.

\begin{table}[t]
\centering
\small
\caption{\textbf{Association transport over distance.}
Recall for locally created associations evaluated at near and far
separations.}
\label{tab:app-distance}
\begin{tabular}{lcc}
\toprule
\textbf{Method} & \textbf{Near} & \textbf{Far} \\
\midrule
DT & \textbf{.859} & \textbf{.827} \\
Sliding-window attention & .142 & .019 \\
\bottomrule
\end{tabular}
\end{table}

DT changes only modestly from \(0.859\) to \(0.827\), whereas local attention
falls from \(0.142\) to \(0.019\). Thus the local operator creates the
cross-position relation, while persistence makes it available beyond the
local receptive field.


\subsection{Long-Context Genomic Extrapolation}
\label{app:long-context}

We next evaluate whether the repaired memory remains useful when sequence
length grows beyond the \(32\,\mathrm{kb}\) context used during pretraining.
The experiment uses TAD-boundary discrimination because the same central
prediction can be evaluated while progressively revealing more surrounding
sequence.

\paragraph{Boundary annotations.}
Human TAD boundaries are derived from the GM12878 Arrowhead domain calls of
Rao et al.~\citep{rao2014}. Domain coordinates are lifted from hg19 to
hg38. A domain is retained only when both endpoints map successfully to the
same target chromosome and the lifted domain length remains within \(20\%\)
of its original length. Shared endpoints between adjacent domains are
collapsed into a single boundary.

\paragraph{Evaluation chromosomes and windows.}
Human evaluation is restricted to chromosomes 8 and 9, which are excluded
from pretraining. For context length \(L\), positive examples are centered on
annotated TAD boundaries. Negative examples are centered at positions at least
\(100\,\mathrm{kb}\) from the nearest annotated boundary. Windows must fit
inside the reference sequence and contain at most \(20\%\) \texttt{N} bases.
The two classes are balanced.

This construction intentionally measures discrimination between
boundary-centered windows and clearly non-boundary controls. It should not be
interpreted as calibrated genome-wide boundary calling, since positions at
intermediate distances from an annotated boundary are not represented by
either extreme.

\paragraph{Frozen representation and probe.}
All pretrained encoder parameters remain frozen. Given hidden states
\(H\in\mathbb{R}^{L\times768}\), we retain the central \(512\) positions and
compute mean and max pooling along the sequence dimension. Concatenating the
two summaries gives a \(1536\)-dimensional representation.

Features are standardized using statistics from the training split. A single
linear logistic probe is optimized with binary cross-entropy for \(400\)
full-batch Adam steps using learning rate \(0.1\) and weight decay
\(10^{-3}\). Performance is reported using AUROC. Public long-context models
use the same window construction, pooling rule, and downstream probe.

\paragraph{Context sweep.}
We evaluate sequence lengths from \(2\,\mathrm{kb}\) to approximately
\(1\,\mathrm{Mb}\). DT parameters are not retrained as context grows.
Consequently, every evaluation above \(32\,\mathrm{kb}\) tests inference-time
extrapolation beyond the sequence length observed during pretraining.

\begin{table*}[t]
\centering
\small
\caption{\textbf{Full TAD-boundary context sweep.}
AUROC of frozen encoder features under the common logistic-probe protocol.
``cap'' denotes contexts beyond a model's native maximum length.}
\label{tab:app-context-full}
\setlength{\tabcolsep}{4.3pt}
\begin{tabular}{lcccccccc}
\toprule
\textbf{Encoder}
& \textbf{2k} & \textbf{8k} & \textbf{32k} & \textbf{65k}
& \textbf{131k} & \textbf{262k} & \textbf{524k} & \textbf{999k} \\
\midrule
\multicolumn{9}{l}{\textit{Repaired memory}} \\
DT + gain 4
& .693 & .760 & .847 & .897 & \textbf{.927} & \textbf{.903}
& \textbf{.870} & \textbf{.856} \\
DT, no gain
& .723 & .771 & .788 & .802 & .834 & .830 & .846 & .722 \\
\midrule

\multicolumn{9}{l}{\textit{Matched architecture / objective controls}} \\
Stock memory
& .716 & .758 & .721 & .749 & .742 & .752 & .750 & .741 \\
Span objective
& .680 & .618 & .703 & .725 & .721 & .762 & .740 & .715 \\
Distal objective
& .659 & .561 & .653 & .629 & .618 & .620 & .640 & .648 \\
\midrule

\multicolumn{9}{l}{\textit{Public long-context encoders}} \\
HyenaDNA-medium
& .678 & .727 & .613 & .717 & .692 & cap & cap & cap \\
HyenaDNA-large
& .667 & .749 & .680 & .686 & .703 & .701 & .720 & .717 \\
Caduceus-PH
& .631 & .771 & .710 & .772 & .810 & cap & cap & cap \\
\bottomrule
\label{app:context-sweep}
\end{tabular}
\end{table*}

\subsection{Selective Genomic Transport}
\label{app:selective-transport}

The context-scaling experiment establishes that distant sequence becomes
decodable. The following experiments ask what information is preferentially
represented and transported. We progressively move from downstream
variant-effect prediction to direct sequence perturbation, selective
long-range transport, real structural variants, and functional constraint.

\subsubsection{Variant-Effect and Long-Range Identity Tasks}
\label{app:variant-effect}

\paragraph{eQTL variant effect.}
The eQTL task asks whether replacing one allele changes regulatory state.
DT is evaluated as a frozen encoder with a lightweight logistic/linear
adapter. Whole blood provides the protocol-matched comparison with frozen
HyenaDNA. We additionally evaluate the same frozen DT representation across
nine GTEx tissues:
cultured fibroblasts (CCF), whole blood (WB), thyroid (Thy),
non-sun-exposed skin (SNSE), sun-exposed lower-leg skin (SSEL),
skeletal muscle (Musc), tibial nerve (Nerv), tibial artery (Art), and
subcutaneous adipose tissue (Adip).

Task-specific expert, CNN, fine-tuned HyenaDNA, and fine-tuned Caduceus
results are taken from the corresponding published benchmark protocols.
Because those models use their own fine-tuning procedures, they provide
external reference points rather than causal comparisons with DT. The
whole-blood frozen-DT versus frozen-HyenaDNA result is the protocol-matched
encoder comparison.

\paragraph{Enhancer--target-gene assignment.}
We pair the variant-effect task with enhancer--target-gene assignment. Here,
the readout must preserve the identity of a particular distant enhancer
associated with a target gene. We use the same frozen-representation
evaluation principle and report AUROC. This complementary task separates
change-sensitive representation from preservation of precise distant-element
identity.


\subsubsection{Boundary Deletion}
\label{app:boundary-deletion}

We next replace observational downstream decoding with direct sequence
perturbation. For \(100\) annotated TAD boundaries evaluated at
\(262\,\mathrm{kb}\) context, we compare the intact sequence with versions in
which \(5\), \(20\), or \(50\,\mathrm{kb}\) centered regions are deleted.

The intact windows have predicted boundary probability
\(0.999\pm0.001\). For each deletion, we report the change relative to the
corresponding intact example.

The largest change occurs for the centered \(5\,\mathrm{kb}\) deletion, while
larger deletions have similar effects. This provides a direct perturbational
test that the long-range readout responds to sequence at the boundary scale.


\subsubsection{Selective Long-Range Transport}
\label{app:selectivity-protocol}

Responding to a boundary does not imply that the memory preferentially
transports boundary information. We therefore fix a distal readout location
and perturb sequence outside its local attention receptive field. For each
example, a boundary-centered perturbation is paired with a matched random
perturbation, and we compute

\begin{equation}
S_{\mathrm{sel}}
=
\frac{\Delta_{\mathrm{boundary}}}
     {\Delta_{\mathrm{random}}},
\label{eq:app-selectivity}
\end{equation}

where \(S_{\mathrm{sel}}>1\) indicates that the boundary perturbation produces
the larger distal effect.

\paragraph{Associative transport over distance.}
We first use the standard associative write, retaining the capacity and binding
repairs but not the change-point target.

\begin{table}[t]
\centering
\small
\caption{\textbf{Selectivity of associative transport.}
Boundary-to-random effect ratio as perturbation distance increases.}
\label{tab:app-associative-selectivity}
\begin{tabular}{lccc}
\toprule
\textbf{Distance} & \textbf{16 kb} & \textbf{32 kb} & \textbf{65 kb} \\
\midrule
Boundary / random & .99 & 1.07 & 1.07 \\
\bottomrule
\end{tabular}
\end{table}

The ratio remains close to one across the full range, establishing that the
repaired associative memory can carry distant effects without strongly
preferring the boundary perturbation.

\paragraph{Change-point write.}
We then keep the repaired memory fixed and replace the associative target with
the bidirectional change-point target defined in
Sec.~\ref{subsec:DT-selectivity}. The comparison is first performed on
controlled boundary examples and then repeated on real gnomAD structural
variants.

\begin{table}[t]
\centering
\small
\caption{\textbf{Selectivity introduced by the change-point write.}
Boundary-to-random effect ratios on controlled boundaries and real structural
variants.}
\label{tab:app-change-selectivity}
\begin{tabular}{lcc}
\toprule
\textbf{Write} & \textbf{Controlled} & \textbf{gnomAD-SV} \\
\midrule
Associative & 1.06 & 1.01 \\
Change-point & \textbf{1.15} & \textbf{1.09} \\
\bottomrule
\end{tabular}
\end{table}

The change-point write increases the selectivity ratio in both settings,
providing the controlled-to-real progression used in
Fig.~\ref{fig:sv}c.


\subsubsection{Functional Constraint}
\label{app:functional-constraint}

Finally, we ask whether the same preference extends beyond boundary identity
to functional constraint. We compare constrained rare deletions with common
tolerated variants and report the probability that the rare deletion produces
the larger propagated effect,

\[
P(\mathrm{rare}>\mathrm{common}).
\]

\begin{table}[t]
\centering
\small
\caption{\textbf{Selective transport under functional constraint.}
Probability that a rare constrained deletion produces a larger distal effect
than a common tolerated variant.}
\label{tab:app-functional-constraint}
\begin{tabular}{lc}
\toprule
\textbf{Write} &
\(\mathbf{P(\mathrm{rare}>\mathrm{common})}\) \\
\midrule
Associative & .42 \\
Change-point & \textbf{.67} \\
\bottomrule
\end{tabular}
\end{table}

The probability rises from \(0.42\) under the associative write to \(0.67\)
under the change-point write.


\subsection{Cross-Species Transfer}
\label{app:cross-species}

The final experiments evaluate whether a representation pretrained on human
DNA remains useful as the evaluation distribution moves away from the human
training genome. We use two settings: a controlled human-to-mouse transfer
with the downstream task held fixed, followed by a broader species sweep with
increasing evolutionary distance.

\subsubsection{Human-to-Mouse Zero-Shot Transfer}
\label{app:human-mouse}

All encoders are pretrained on human sequence and remain frozen during
evaluation. Human TAD-boundary results use held-out chromosomes 8 and 9.
Mouse evaluation uses mm10 chromosomes 1--5 with no encoder adaptation.
The same TAD-boundary representation and logistic-probe protocol is applied
to both species.

We compare the full DT encoder with the same encoder after disabling its
fast-weight memory path. Caduceus-PS provides a bidirectional state-space
reference with native context up to \(131\,\mathrm{kb}\), while
HyenaDNA-1m provides a native megabase-context reference.

\begin{table*}[t]
\centering
\small
\caption{\textbf{Zero-shot human-to-mouse transfer.}
TAD-boundary AUROC from human-pretrained frozen encoders under the matched
logistic-probe protocol. ``cap'' denotes context beyond a model's native
maximum length.}
\label{tab:app-crossspecies}
\setlength{\tabcolsep}{4pt}
\begin{tabular}{llccccccc}
\toprule
 & \textbf{Encoder}
 & \textbf{2k} & \textbf{8k} & \textbf{32k}
 & \textbf{131k} & \textbf{262k} & \textbf{524k} & \textbf{1M} \\
\midrule

\multirow{4}{*}{\rotatebox{90}{Human}}
& DT, memory on
& .693 & .760 & .847 & .927 & .903 & .870 & .856 \\

& Same encoder, memory off
& .640 & .641 & .617 & .623 & .658 & .625 & .610 \\

& Caduceus-PS
& .668 & .739 & .782 & .827 & cap & cap & cap \\

& HyenaDNA-1m
& .633 & .701 & .640 & .660 & .776 & .712 & .784 \\

\midrule

\multirow{4}{*}{\rotatebox{90}{Mouse}}
& DT, memory on
& .677 & .699 & .786 & .785 & .761 & .783 & .796 \\

& Same encoder, memory off
& .653 & .583 & .602 & .639 & .630 & .648 & .713 \\

& Caduceus-PS
& .710 & .697 & .752 & .773 & cap & cap & cap \\

& HyenaDNA-1m
& .615 & .711 & .693 & .733 & .713 & .698 & .710 \\

\bottomrule
\end{tabular}
\end{table*}

The human-to-mouse experiment changes the genome while retaining the task,
probe, and human-pretrained encoder, providing the most controlled
cross-species test in the paper.

\subsubsection{Increasing Evolutionary Shift}
\label{app:evolutionary-shift}

We next broaden the shift beyond a single target species. Starting from human,
we evaluate chimpanzee, macaque, mouse, and chicken, ordered by increasing
evolutionary distance from the human training genome. For each species, we
compare the frozen memory state with the corresponding test-time-adapted state.

Frozen and adapted values remain closely aligned across the mammalian species
in this sweep, while chicken provides the first non-mammalian evaluation and
departs from the mammalian pattern.

\end{document}